\documentclass{article} 

\usepackage{hyperref}
\usepackage{url}

\usepackage{amssymb}
\usepackage{amsmath}
\usepackage{multirow}

\usepackage{graphicx}

\usepackage{icml2014}
\icmltitlerunning{Predict Information Diffusion using a Latent Representation Space}

\begin{document} 

\twocolumn[
\icmltitle{Predict Information Diffusion using a Latent Representation Space}

\icmlauthor{C\'edric Lagnier}{cedric.lagnier@lip6.fr}
\icmlauthor{Simon Bourigault}{simon.bouraigault@lip6.fr}
\icmlauthor{Sylvain Lamprier}{sylvain.lamprier@lip6.fr}
\icmlauthor{Ludovic Denoyer}{ludovic.denoyer@lip6.fr}
\icmlauthor{Patrick Gallinari}{patrick.gallinari@lip6.fr}
\icmladdress{Sorbonne Universites, UPMC Univ Paris 06, UMR 7606, LIP6, F-75005, Paris, France\\
			CNRS, UMR 7606, LIP6, F-75005, Paris, France}

\icmlkeywords{machine learning, information diffusion, social networks}

\vskip 0.3in
]

\begin{abstract}
Information propagation is a hard task where the goal is to predict users behavior.
We introduce an extension of a model which make use of a kernel to modelize diffusion in
a latent space. This extension introduce a threhsold to differentiate if users are
contaminated or not.
\end{abstract}


\section{Introduction}

The emergence of Social Networks and Social Media sites has motivated a large amount of recent research. Different generic tasks, such
as Social Network Analysis, Social Network annotation, Community Detection or Link Prediction, are currently studied. Another active
research topic is the study of temporal propagation of information through this type of media. It aims at studying how interactions
between users, like sharing a link on facebook or retweeting something on Twitter, effects the spread of items such as pictures, videos
or gossip on the internet.
While the study of this word-of-mouth phenomenon pre-dates the development of computer science, the amount of data made available
by the growth of online social networks offers an unprecedented field of study and enabled new developments.
Propagation models aim at predicting and understanding the dynamic of observed propagation.


In this paper, we propose a new diffusion model based on the
heat diffusion. It
aims to project users in a latent space where propagation occurs
like the heat diffusion. This projection is based on the order in which
users have been infected in cascades of the train dataset. In order to
be able to find which users have been infected and not only who is
the most likely to be infected, we define a threshold to split users
in two groups: infected or not.
This model is an extension of the CDK model (Content Diffusion Kernel)
presented in \cite{bourigault-2008} where no threshold was defined.


\section{Notations}

Traditionally, diffusion on networks is represented with the notion of \emph{cascade}. A cascade is a sequence of users infected by
some information (for instance, it could be the list of users who "liked" a specific YouTube video). A cascade describes to whom and
when an item spreads through the network, but not \emph{how} diffusion happens: while it is easy to know \emph{when} a user got infected
by some content, it is usually not possible to know \emph{who} infected him.

Given a social network composed of a set of $N$ users $\mathcal{U}=(u_1,....,u_N)$, cascades correspond to sets of users infected by
the propagated information. Depending on the kind of network and the task in concern, the propagated information can for instance
correspond to a given topic, a particular url, a specific tag, etc...
In the following, we consider $\mathcal{C}$ as a set of cascades over a given network, and two sets of distinct cascades:
$\mathcal{C}_\ell \subseteq \mathcal{C}$ the set of training cascades and $\mathcal{C}_t \subseteq \mathcal{C}$ the set of testing cascades.
A cascade $c \in \mathcal{C}$ is defined as:
\begin{itemize}
\item A source $s^c \in \mathcal{U}$ which is the user at the source of the cascade - i.e, the first user that published the item concerned by the diffusion.
\item A set of contaminated users $S^c \subset \mathcal{U}$ such that $u_i \in S^c$ means that $u_i$ has participated to the cascade $c$
$\bar{S^c}$ is the set of users who have not participated in $c$.
\item A contamination timestamp function defined over $S^c$ such that $t^c(u_i)$ corresponds to the timestamp at which $u_i \in S^c$ has first participated in the cascade. We consider that the contamination timestamp of the source is equal to $0$.
\end{itemize}


\section{Model}

The proposed model aims at predicting information diffusion. The central idea of this model is to map the observed information
diffusion process into a heat diffusion process in a continuous (euclidean) space. To perform this, we learn diffusion kernels that
capture the dynamics of diffusion from a set of training cascades. Let us denote $\mathcal{Z} = \mathbb{R}^n$ an euclidean space of
dimension $n$ - also called \textit{latent space}. Learning such a diffusion kernel comes down in our case to learning a
mapping of each node of the network to
a particular location in $\mathcal{Z}$ such that, for a given metric, the latent space explains the contamination timestamps
observed in the training cascades.

\paragraph{Learning using a diffusion kernel}

We define a diffusion kernel $K(t,y,x)$ such that $K :
\mathbb{R}^{+} \times \mathcal{X} \times \mathcal{X} \rightarrow \mathbb{R}$
which models the heat diffusion in a latent space. It corresponds here to the
contamination propensity of a node \emph{x} at time $t$ given a particular information source \emph{y}. For learning the kernel function, there is however no full supervision available - this would correspond to a continuous time function giving the heat evolution at any point. The observations only provide the contamination time of the different nodes in a cascade. This partial supervision will be used to constrain the kernel to contaminate the different nodes \textbf{in their actual temporal order of infection}.

In practice, we will use the following constraints:
\begin{itemize}
\item Given two nodes $u_i$ and $u_j$ such that $u_i$ and $u_j$ are contaminated during cascade $c$ - i.e $u_i \in S^c$ and $u_j \in S^c$ - and respecting $t^c(u_i) < t^c(u_j)$, $K_Z$ must be defined such that $\forall t, K_Z(t,s^c,u_i) > K_Z(t,s^c,u_j)$
\item We define a heat threshold $h_{\tau}$ which determine the
heat users have to reach to be contaminated. Thus:
  \begin{itemize}
  \item Given a node $u_i$ such that $u_i$ is contaminated during a
  cascade $c$, $K_Z$ must be defined such that
  $\exists t, K_Z(t,s^c,u_i) > h_{\tau}$
  \item Given a node $u_i$ such that $u_i$ is not contaminated during a
  cascade $c$, $K_Z$ must be defined such that
  $\forall t, K_Z(t,s^c,u_i) < h_{\tau}$
  \end{itemize}
\end{itemize}

These constraints basically aim at finding embeddings such that users who are contaminated first are closer to the source of the contamination than users contaminated later (or not contaminated at all). $h_{\tau}$ is a unique heat threshold which split users in two groups in order to determine which users will be contaminated and not only an order of contamination.
Based on the heat equation, we can thus easily rewrite
these three constraints as:
\begin{equation}
\label{embedded_constraints}
\begin{aligned}
&\forall u_i \in S^c, \quad \quad ||z_{s^c} - z_{u_i} ||^2 < \tau & \\
&\forall u_i \in \bar{S^c}, \quad \quad \tau < ||z_{s^c} - z_{u_i} ||^2 & \\
&\forall (u_i,u_j) \in S^c \times S^c\\
&\quad \quad t^c(u_i) < t^c(u_j) \Rightarrow  ||z_{s^c} - z_{u_i} ||^2 < ||z_{s^c} - z_{u_j} ||^2 & \\
\end{aligned}
\end{equation}
where $\tau$ is a distance threshold. It correponds to the distance from
the source of the diffusion beyond which users are not contaminated:
their heat never reach $h_{\tau}$.

By the use of classical hinge loss functions, these constraints can be handled by defining a ranking objective $\Delta_{rank}$ such as:
\begin{equation}
\label{hinge}
\begin{aligned}
& \Delta_{rank}(K_Z(.,s^c,.),c,\tau) = \\
&\sum\limits_{u_i \in S^c} max(0, 1-(\tau - ||z_{s^c} - z_{u_i} ||^2))\\
&+\sum\limits_{u_i \in \bar{S^c}} max(0, 1-(||z_{s^c} - z_{u_i} ||^2 - \tau)) \\
&+\displaystyle\sum_{u_i,u_j \in S^c \times S^c \atop t^c(u_i) < t^c(u_j)} max(0, 1-(||z_{s^c} - z_{u_j} ||^2 - ||z_{s^c} - z_{u_i} ||^2))
\end{aligned}
\end{equation}

\paragraph{Learning Algorithm}

The final training objective is:
\begin{equation}
\mathcal{L}_{rank}(Z,\tau) =  \sum\limits_{c \in \mathcal{C}_\ell}
\Delta_{rank}(K_Z(.,s^c,.),c,\tau)
\end{equation}
We name this model "Content Diffusion Kernel with Threshold" (CDKT). Different methods can be used to optimize the objective function. We
propose to use a classical stochastic gradient descent method, which iterates until having
reached a stop criterion (typically a number of iterations
without significant improvement of the global loss). After having randomly initialized\footnote{Different initialization strategies
can be adopted. In our experiments, we used an uniform initialization between -1 and 1.} all embeddings for nodes in $\mathcal{U}$,
the algorithm samples at each iteration a cascade $c$ from the training set $\mathcal{C}_\ell$ and two nodes $u_i$ and $u_j$
with $u_j$ a node that is either non-infected, or contaminated after $u_i$ in the diffusion process described by cascade $c$.
If constraints defined in
equation \ref{embedded_constraints} are not respected with a sufficient margin\footnote{As defined by the hinge loss function,
see equation \ref{hinge}.} for this cascade $c$ and the nodes $u_i$ and $u_j$, embeddings $z_{u_i}$, $z_{u_j}$, $z_{s^c}$ and $\tau$
are modified towards their respective steepest gradient direction with a learning rate $\alpha$  which is a decreasing
function of the number of iterations.
The learning process is illustrated in algorithm \ref{alg:sgd1}.

\begin{algorithm}[t]
\begin{algorithmic}[1]
\Procedure{SGD Rank Diffusion Kernel Learning}{}
\State $t \gets 0$
\State $\forall u \in \mathcal{U}, z_u^{(t)} \gets random$
\State $\tau^{(t)} \gets random$
\While{$t < T $ }
	\State Sample $c \in \mathcal{C}_\ell$
	\State Sample $u_i \in S^c$
	\State Sample $u_j \in \mathcal{U}  \text{ , } t^c(u_i) < t^c(u_j)$ or $u_j \in \bar{S^c}$ 
	\State $d_i \gets ||z_{s^c}^{(t)} - z_{u_i}^{(t)} ||^2$
	\State $d_j \gets ||z_{s^c}^{(t)} - z_{u_j}^{(t)} ||^2$
	\If{$u_j \in S^c$}
		\If{$(d_j-d_i) < 1$}
			\State $z_{u_i}^{(t+1)} \gets z_{u_i}^{(t)} + \alpha(t) \times 2(z_{s^c}^{(t)}-z_{u_i}^{(t)})$
			\State $z_{u_j}^{(t+1)} \gets z_{u_j}^{(t)} + \alpha(t) \times 2(z_{u_j}^{(t)}-z_{s^c}^{(t)})$
			\State $z_{s^c}^{(t+1)} \gets z_{s^c}^{(t)} + \alpha(t) \times 2(z_{u_i}^{(t)}-z_{u_j}^{(t)})$
		\EndIf
	\EndIf
	\If{$(\tau^{(t)}-d_i) < 1$}
		\State $z_{u_i}^{(t+1)} \gets z_{u_i}^{(t)} + \alpha(t) \times 2(z_{s^c}^{(t)}-z_{u_i}^{(t)})$
		\State $\tau^{(t+1)} \gets \tau^{(t)} - \alpha(t)$
	\EndIf
	\If{$(d_j-\tau^{(t)}) < 1$}
		\State $z_{u_j}^{(t+1)} \gets z_{u_j}^{(t)} + \alpha(t) \times 2(z_{u_j}^{(t)}-z_{s^c}^{(t)})$
		\State $\tau^{(t+1)} \gets \tau^{(t)} + \alpha(t)$
	\EndIf
	\State $t \gets t+1$
\EndWhile
\State $Z \gets Z^{(t)}$
\EndProcedure
\end{algorithmic}
\caption{Stochastic gradient descent algorithm}
\label{alg:sgd1}
\end{algorithm}




\section{Experimentations}

\paragraph{Datasets}

We tested our model on several datasets from various online sources: \emph{ICWSM} \cite{icwsmdata}, \emph{Memetracker} \cite{memetracker} and \emph{Digg}.
The first two datasets are sets of blog posts crawled from the web. We define a cascade as a
set of posts linked together by hyperlinks. \emph{Digg} est a plateform where users can share
news stories with each other. A cascade is thus the set of users who have share the same story.
We filtered the users of each dataset to keep about 5000 users with the most posts.

\paragraph{Quality of the ranking}

\begin{table}
\center

\begin{tabular}{|c|c|c|c|}
\hline 
Model & Memetracker & ICWSM & Digg\tabularnewline
\hline 
\hline 
CDK-500 &0.363&0.773&0.280\tabularnewline
\hline
CDKT-500 & 0.324 & 0.746 & 0.233 \tabularnewline
\hline
IC & 0.372 & 0.712 & 0.197 \tabularnewline
\hline
Netrate & 0.287 & 0.187 & 0.162 \tabularnewline
\hline
Heat Diff. & 0.374 & 0.483 & 0.082 \tabularnewline
\hline
\end{tabular}

\caption{MAP on $3$ real datasets: \emph{Memetracker}, \emph{ICWSM} and \emph{Digg}.  Results of CDK and CDKT are given for a latent space of 500 dimensions.\label{tableReal}}

\end{table}

In order to test the quality of this model, we compared it to several baselines using the
same protocol we used in \cite{bourigault-2008}. The goal is to compute
the average precision the model obtains on all cascades.
We show the results of 3 baselines IC \cite{saito-2008}, Netrate \cite{gomez-2011} and Heat Diffusion \cite{ma-2008} and the 2 latent models
CDK and CDKT. IC obtains better results than other baselines.

CDK obtains slightly better results than CDKT. They outperform baselines on both \emph{ICWSM} and \emph{Digg} while IC obtains better results on \emph{Memetracker}.

\paragraph{Learning and Inference complexity}

Let $T$ be the number of iterations. The learning complexity is $O(T \times n)$, where $n$ is the size of the latent space.  
Once $Z$ has been learned, the inference process is simple. For a cascade $c$, we just compute the distance between the user $s^c$ and every other user in $\mathcal{U}$. The inference complexity for every cascade is then $O(N \times n)$, where $N$ is the number of users. Considering that $n \ll N$, this turns out to be much smaller than the complexity of most alternative discrete methods.  
For instance, the inference step of the very famous
Independant Cascade model(IC), which 
is a probabilistic model where diffusion propabilities are defined on edges of the network's graph, requires to consider at each time step of the diffusion $t$ every possible infection situation at previous time $t-1$, which quickly becomes untractable.
In practice, inference of graphical models is done by employing a Monte-Carlo approximation that consists in performing a high amount of simulations of the diffusion process starting from the source of the cascade and following the diffusion probabilities on links of the graph.
The inference complexity of this
approximation of IC is $O(r \times \hat{Succs} \times \hat{|S^c|})$, where $\hat{|S^c|}$ is the average number of infected nodes in the performed simulations, $\hat{Succs}$ is their average outdegree and $r$ is the number of simulations used for the MCMC approximation. The weaker the probabilities defined on links are, the greater $r$ must be set to obtain a correct approximation of the distribution of final infection probabilities.





\paragraph{Comparison between CDK and CDKT}

We compare here the two kernel models in the task of predicting which users will be contaminated
at the end of the diffusion. The main problem to achieve this task with the CDK model is
that all cascades are not on the same scale and it is very difficult to find a unique
threshold which properly split data in two clusters: contaminated or not. For this reason we
don't have any threshold for the CDK model and thus couldn't compare clusters made.

We use the following protocol: after predicting a score of contamination for each user for each cascade, we group all
cascades in a unique set. The goal is to see if the models can find users contaminated by
any cascade in this set. If there is a unique threshold, they should be able to do so.





\begin{table}
\center

\begin{tabular}{|c|c|c|c|}
\hline 
 & Memetracker & ICWSM & Digg\tabularnewline
\hline 
\hline 
CDK-50 & 0.0001 & 1.0 & 0.450\tabularnewline
\hline
CDKT-50 & 0.297 & 1.0 & 0.543\tabularnewline
\hline
\hline 
CDK-500 & 0.625 & 0.794 & 0.067\tabularnewline
\hline
CDKT-500 & 0.417 & 1.0 & 0.862\tabularnewline
\hline
\end{tabular}

\caption{P@50 on $3$ real datasets: \emph{Memetracker}, \emph{ICWSM} and \emph{Digg}. Results of CDK and CDKT are given for two values of $n$, the dimension of the latent space $\mathcal{Z}$ (50 and 500 dimensions).\label{tabp50}}

\end{table}

Table \ref{tabp50} shows the precision at rank 50 (P@50) on all datasets for the two models using
two latent spaces with different dimension. It corresponds to
their ability to find 50 contaminated users.
We see that both CDK and CDKT obtain better results on \emph{ICWSM}.
They also obtain a better MAP. This is because the \emph{ICWSM} dataset
is easier than the 2 others.
While CDKT is obtain better resultats than CDK on most of the datasets
and dimension spaces.

\begin{figure}
\center
\includegraphics[width=0.95\linewidth]{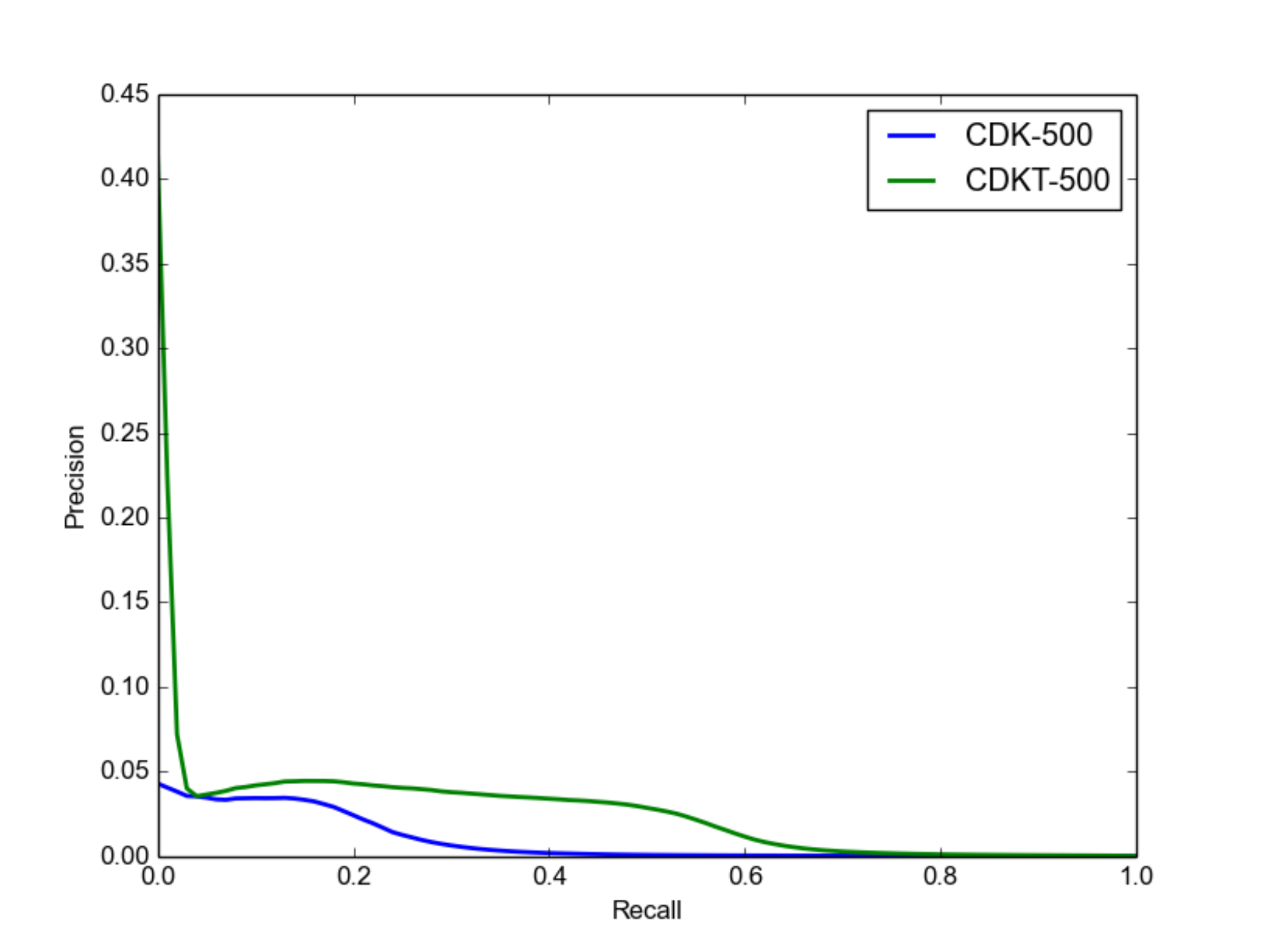}
\caption{Precision/Recall curve for models using a latent space of 500 dimensions on
the \emph{Digg} dataset.\label{curve}}
\end{figure}

As the P@50 doesn't show all information, figure \ref{curve} shows the precision/recall curve
for the \emph{Digg} datasets in 500 dimensions spaces. As for the P@50,
the curve shows that CDKT is better than CDK.


\section{Conclusion}

The CDK model use the phenomenon of heat diffusion to modelize the propagation of content
in a latent space. This model is based on a ranking of users and because of the different scale
of each representation, there is no easy way to find which user will be contaminated.
We proposed in this article an extension of this model CDKT which learns a threshold to split
users in two groups: contaminated or not. On several real datasets, we showed that this
model is better to find contaminated users.
Our next step with this model will be to understand in which contexts (low/hight diffusion, network type, etc) it outperfoms CDK.




\bibliography{article.bib}
\bibliographystyle{icml2014}










\end{document}